\documentclass[10pt, a4paper]{article}

\usepackage[most]{tcolorbox} 
\usepackage{caption}

\newtcolorbox{promptbox}{
    colback=white,      
    colframe=black,     
    arc=0pt,            
    outer arc=0pt,
    boxrule=0.5pt,      
    width=\textwidth,   
    enhanced,
    breakable           
}

\usepackage{lrec2026} 
\usepackage{tabularx} 
\usepackage{booktabs} 
\usepackage{graphicx}
\usepackage{subfig}
\usepackage{multirow}
\usepackage{geometry} 
\usepackage{makecell} 

\title{Beyond Transcripts: Iterative Peer-Editing with Audio Unlocks High-Quality Human Summaries of Conversational Speech}

\name{{\large Kaavya Chaparala, Thomas Thebaud, Jesus Villalba Lopez,}\\ {\large \textbf{ Laureano Moro-Velazquez, Peter Viechnicki and Najim Dehak}}}

\address{Johns Hopkins University\\
         3400 N. Charles St., Baltimore, MD 21218\\
         kchapar1@jh.edu} 
\abstract{There are not enough established benchmarks for the task fo speech summarization. Creating new benchmarks demands human annotation, as LLMs could embed systemic errors and bias into datasets. We test ten annotation workflows varying input modality (audio, transcript, or both) and the inclusion of editing (self or peer-editing) to investigate potential quality tradeoffs from using human annotators to summarize audio. We compare human audio-based summaries to human transcript-based summaries to track the impact of the different information modalities on summary quality. We also compare the human outputs against four LLM benchmarks (three text, one audio) to examine whether human-written summaries are less informative than highly fluent automated outputs. We find that audio-based summaries are less informative and more compressed than transcript summaries. However, iterative peer-editing with audio mitigates this difference, enabling audio-based summaries to be as informative as their transcript counterparts and LLM summaries. These findings validate iterative peer-editing among human annotators for the creation of benchmarks informed by both lexical and prosodic information. This enables crucial dataset collection even in setting where transcripts are unavailable.
 \\ \newline \Keywords{speech summarization, human annotation, abstractive summarization, spoken dialogue} }

\begin{document}

\maketitleabstract
\section{Introduction} 

The field of conversational speech summarization relies heavily on benchmark datasets like AMI and ICSI \citep{McCowan:2005:AMI,Janin:2003:ICSI}. These benchmarks center on formal meeting settings, and the field requires equally high-quality benchmarks for unstructured, casual settings. 

These essential benchmarks should be created with human annotators rather than automated methods. While Large Language Models (LLMs) are powerful tools, their outputs are prone to systemic errors, hallucinations, and are highly variable according to the input prompt \citep{lu2025systematicbiaslargelanguage,belem-mds_hallucination,fu2024largelanguagemodelsllms}. If embedded into a benchmark dataset, LLM summaries could steer the field of speech summarization to replicate these undesirable behaviors. This is especially true in the nascent domain of Audio LLMs, where the unique failures are far less understood than text-based LLMs. By contrast, human annotators can process a speech signal without need for intermediate transcriptions, and the errors they make are varied and non-systemic \citep{reinhart-llm-writing, maynez-etal-2023-benchmarking}. Using a team of annotators ensures that individual errors average out across the dataset, yielding a more robust and reliable ground truth for system evaluation \citep{baumann2025largelanguagemodelhacking}.

However, using humans over LLM may incur certain tradeoffs in quality. Previous research has shown that human evaluators often express a strong preference for LLM-written summaries due to their superior fluency and coherence \citep{Pu-2023, herbold-2023}. This suggests human-written summaries might be of poorer overall quality (less informative or less accurate) than their automated counterparts. A second potential trade-off arises from the input modality itself: the temporal nature of audio may exert a higher cognitive load on annotators compared to a transcript which can be scanned visually to reference information \citet{Sharma:2024:SpeechvsText}. This increased load may hinder information extraction and result in summaries which incorporate prosodic cues but which are less informative overall.

To address these challenges, we design a  study to identify the human workflow that maximizes summary quality without sacrificing the inclusion of prosodic context. Our research is guided by two hypotheses:

\begin{itemize}
\item Audio-based summaries will be less informative than transcript-based summaries

\item With the correct workflow, human annotators can produce speech summaries that are just as informative as LLM-generated summaries.
\end{itemize}

We design and explore ten annotation workflows, varying the input modality (audio, transcript, or both) and inclusion of editing (self-editing, peer-editing, and iterative peer-editing). We compare the human-written summaries against each other to validate the first hypothesis. And we evaluate these outputs against summaries generated by four LLMs (three text-based LLMs and one audio LLM) to validate the second hypothesis.

This work provides the first systematic study of human conversational speech summarization workflows and yields three key contributions to the field:

\begin{itemize}
    \item We provide nuanced confirmation of our first hypothesis. We demonstrate that audio-based summaries that have not undergone any editing are shorter and less informative than transcript-based summarization.
    
    \item We identify iterative peer-editing as the most effective methodology to overcome the differences between audio-based and transcript-based summaries and produce audio-based summaries as informative as LLM summaries.  

    \item We show that human-written summaries are as informative as LLM summaries when the human summaries have undergone iterative peer-editing. 
\end{itemize}

\section{Related Work}

Below, we review related work on editing workflows for writing tasks, the impacts of input modality on summarization, and human vs. LLM-generated summaries.

Research in ESL (English as a Second Language) writing highlights the effects of self-editing versus peer editing to improve writing quality. Peer-editing detects more rule-based errors than self-editing, leads to more thorough revisions, and results in greater improvements to ESL students' writing \citep{mawlawi-2010,Ozogul-2009, Fhaeizdhyall-2018}. In NLP settings, peer-editing and iterative revision also significantly enhance text quality \citep{du-etal-2022-understanding-iterative, bernstein-2015}. While these findings reveal the strengths of peer and iterative editing workflows, the applications of these methods to the complex task of summarization is largely unexplored in NLP.

It is especially important to understand if editing can improve human summaries given the significant content  differences between human-written and LLM-generated summaries. LLMs are perceived by humans as producing more coherent and fluent summaries than human annotators \citep{Pu-2023, herbold-2023}. However, in some contexts human summaries are more faithful to the source as LLMs can include generalizations which  misrepresent details \citep{Peters-2025}. Considering these tradeoffs, we pose the following question: can the application of peer-editing and iterative-editing workflows improve human-written summaries to a point where they rival LLM-generated summaries?   

Another key factor influencing summarization is the input modality, specifically whether the source is spoken audio or written text. Research finds that while general comprehension from both modalities is comparable, reading text allows for deeper understanding of details \citep{kauchak-2019, lisell-2022}. Notably, these studies are performed with content that is native to text such as snippets from Wikipedia articles. Fewer studies look at the effects of input modality when the source is naturally occurring in speech. And to our knowledge, only one study explores how these differences translate into summarization performance. \citet{Sharma:2024:SpeechvsText} compare summaries derived from audio and text but do not consider workflows incorporating self- and peer-editing, iterative editing, or the inclusion of both modalities simultaneously. This gap motivates the present exploration of workflow designs that account for modality-specific challenges in summarization.

\section{Methods}
\subsection{Data Preprocessing}
We randomly select four conversations from Switchboard Dialogue Act (SWBDA)  and nine from CallHome \citep{Switchboard,CallHome} to form a set of 13 conversations. 

\begin{figure}
\begin{center}
\includegraphics[width=\columnwidth]{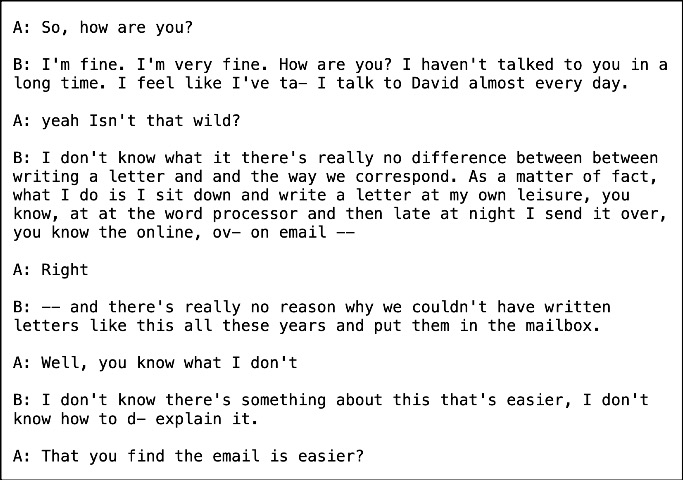}
\caption{Preprocessed conversation transcript from CallHome.}
\label{fig.1}
\end{center}
\end{figure}

Both SWBDA and CallHome are datasets of telephone conversations with audio recordings and manual transcripts. These datasets were selected because the inclusion of manual transcripts allowed us to examine the different effects of text and audio on human summarization without contamination from ASR errors. Conversation length was standardized across all 13 samples by truncating conversations from CallHome to five minutes to match the length of SWBDA. Annotators were informed that the conversations may end abruptly so that truncation would not affect the written summaries.  Transcripts were preprocessed to remove dialogue act tags, leaving clean natural language text as shown in Figure \ref{fig.1}.

\subsection{Annotators} 
We employed 18 U.S. citizens, all native speakers of English, as annotators. We required these qualifications because SWBDA and CallHome both contain an amount of U.S.-centric cultural and linguistic context which non-American or non-native English speakers may find confusing. Annotators are a mix of university students and working professionals, as prior work has shown non-expert annotators produce summaries that are as factually informative as expert annotators \citep{Sharma:2024:SpeechvsText}. Each annotator underwent training: they were given guidelines for writing abstractive summaries and asked to summarize a sample conversation transcript from CallHome or SWBDA. A member of the research team reviewed their summary, identified areas where the writing deviated from the guidelines, and asked the annotator to correct their summary. Annotators then repeated the exercise for different conversations until they had written a summary that adhered to the summarization guidelines. 

\begin{figure*}[!ht]
\begin{center}
\includegraphics[width=\textwidth]{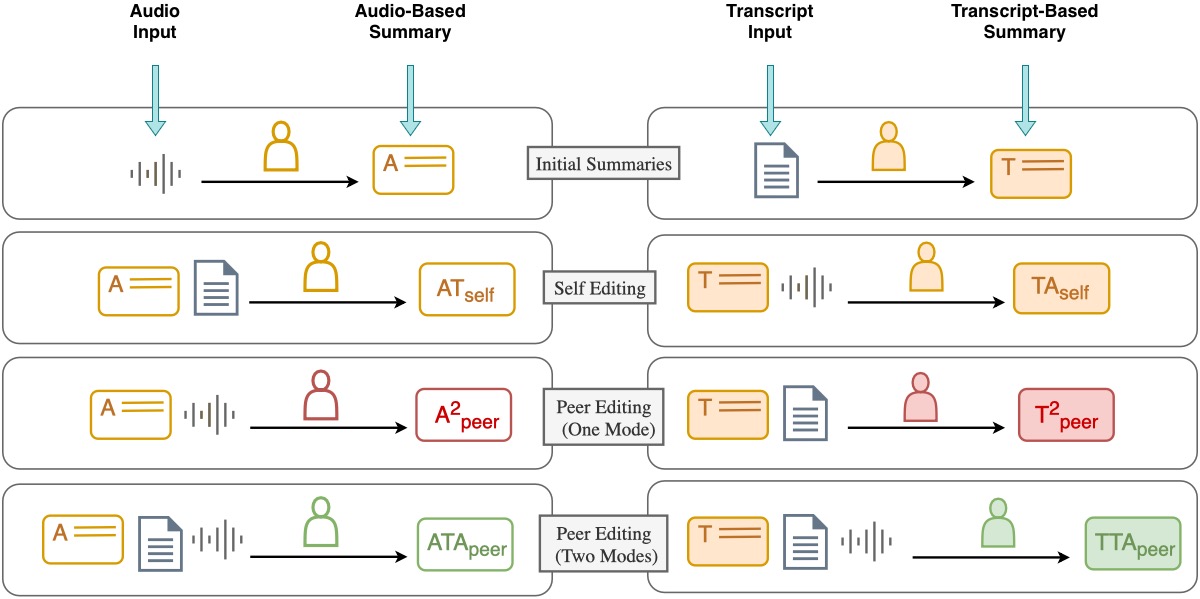}
\caption{Overview of human annotation workflows varying in source representation and editing style (not including iterative peer-editing).}
\label{fig.2}
\end{center}
\end{figure*}

\begin{figure}[!ht]
\begin{center}
\includegraphics[width=\columnwidth]{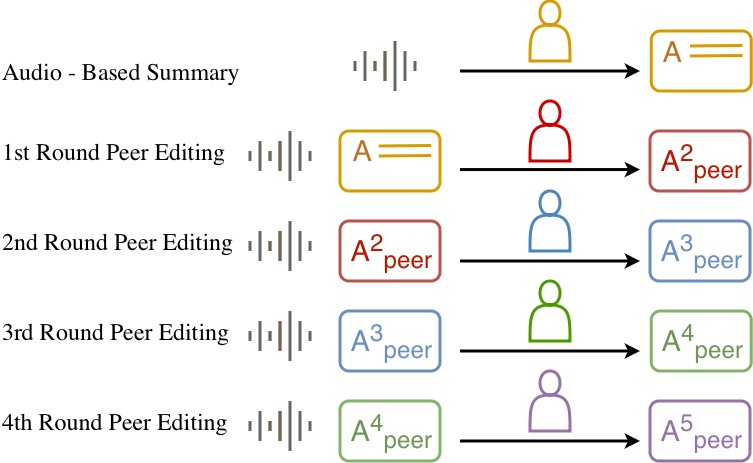}
\caption{Workflow for iterative peer-editing of an audio-based summary.}
\label{fig.3}
\end{center}
\end{figure}

\subsection{Data Annotation}
We designed ten human-annotation workflows to investigate four different factors: the mode of information (audio versus transcript), a single round of editing (by one's self or by a peer), multiple rounds of editing, and the provision of different modalities during editing. This began with the collection of single-modality summaries: annotators either listened to the audio and then wrote an audio-based summary ($A$), or they read the transcript and wrote a transcript-based summary ($T$). 

Next, each annotator was given access to the alternate mode of information, with instructions to read or listen and then edit their initial summary to reflect any changes in their understanding ($AT_{self}$, $TA_{self}$). 
Note that annotators could still interact with the original mode of information when the alternate modality was provided. 

The original audio-based and transcript-based summaries were also given to other annotators to peer-edit. This was done in two ways: the peer-editor could be given both the audio and the transcript to reference when editing ($ATA_{peer}$,$TTA_{peer}$), or they could only be given the same modality of information that the summary was written from ($A^2_{peer}, T^2_{peer}$). Figure \ref{fig.2} depicts these eight workflows for collecting human-written summaries.

In addition to these eight workflows, we introduce two more: iterative peer-editing of the audio- and transcript-based summaries. Iterative peer-editing is comprised of four editing rounds. In each round, a new annotator receives the summary and edits it using the same modality that the original summary was written from ($A^{2-5}_{peer}$, $T^{2-5}_{peer}$). Figure \ref{fig.3} illustrates iterative peer-editing of audio-based summaries.

In every annotation workflow, annotators were able to dynamically interact with whichever mode(s) of information they were given (ie. scroll through the transcript, scrub through the audio). Annotators never worked on the same conversation twice, and the order in which annotators annotated the conversations was randomized.

\subsubsection{LLM Summaries} 
\begin{figure*}[t!]
    \subfloat[\label{genworkflow}]{%
            \includegraphics[trim=5 0 5 0, clip,  width=0.33\textwidth]{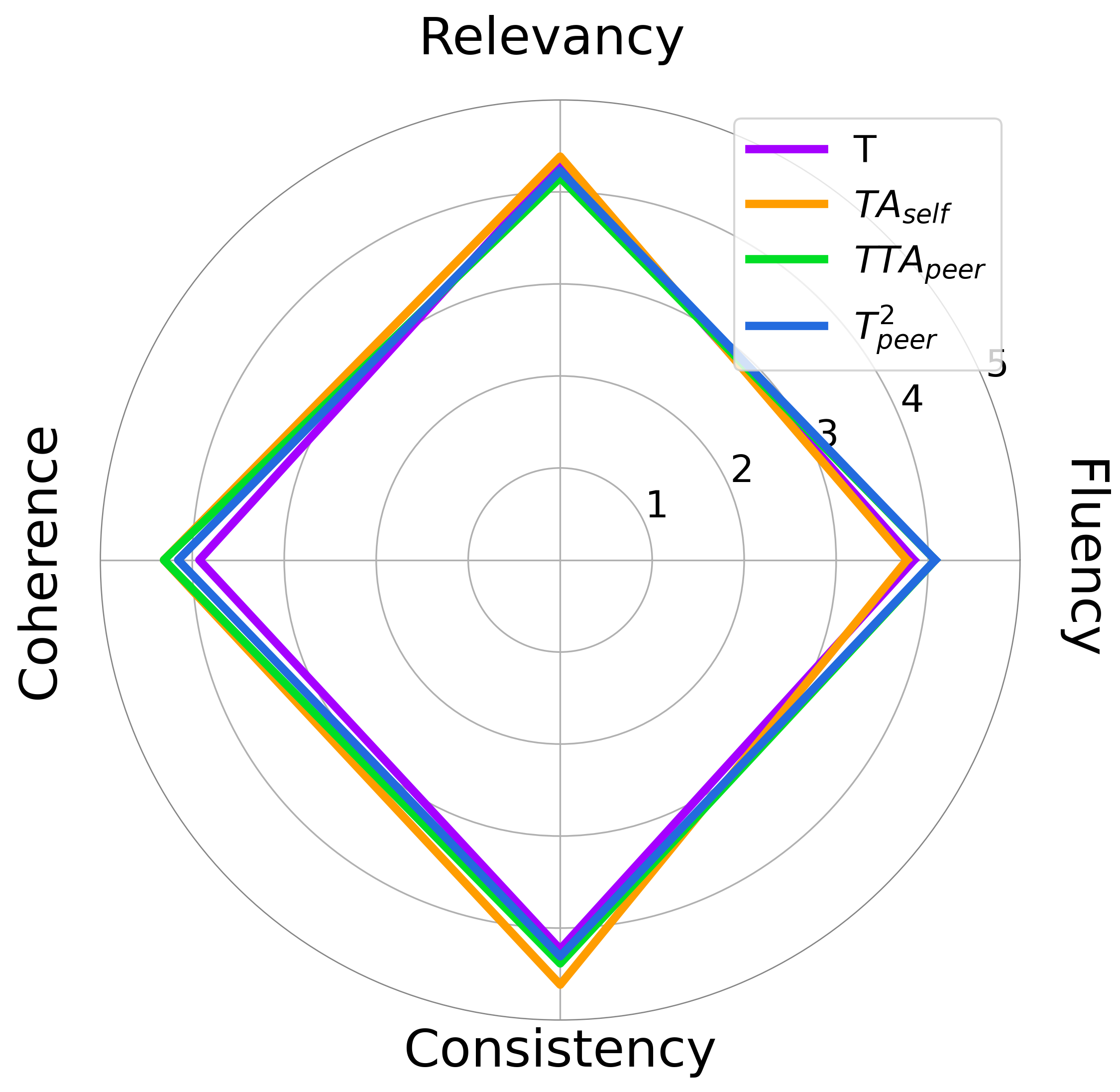}}
\quad
    \subfloat[\label{pyramidprocess} ]{%
    \includegraphics[trim=5 0 5 0, clip, width=0.33\textwidth]{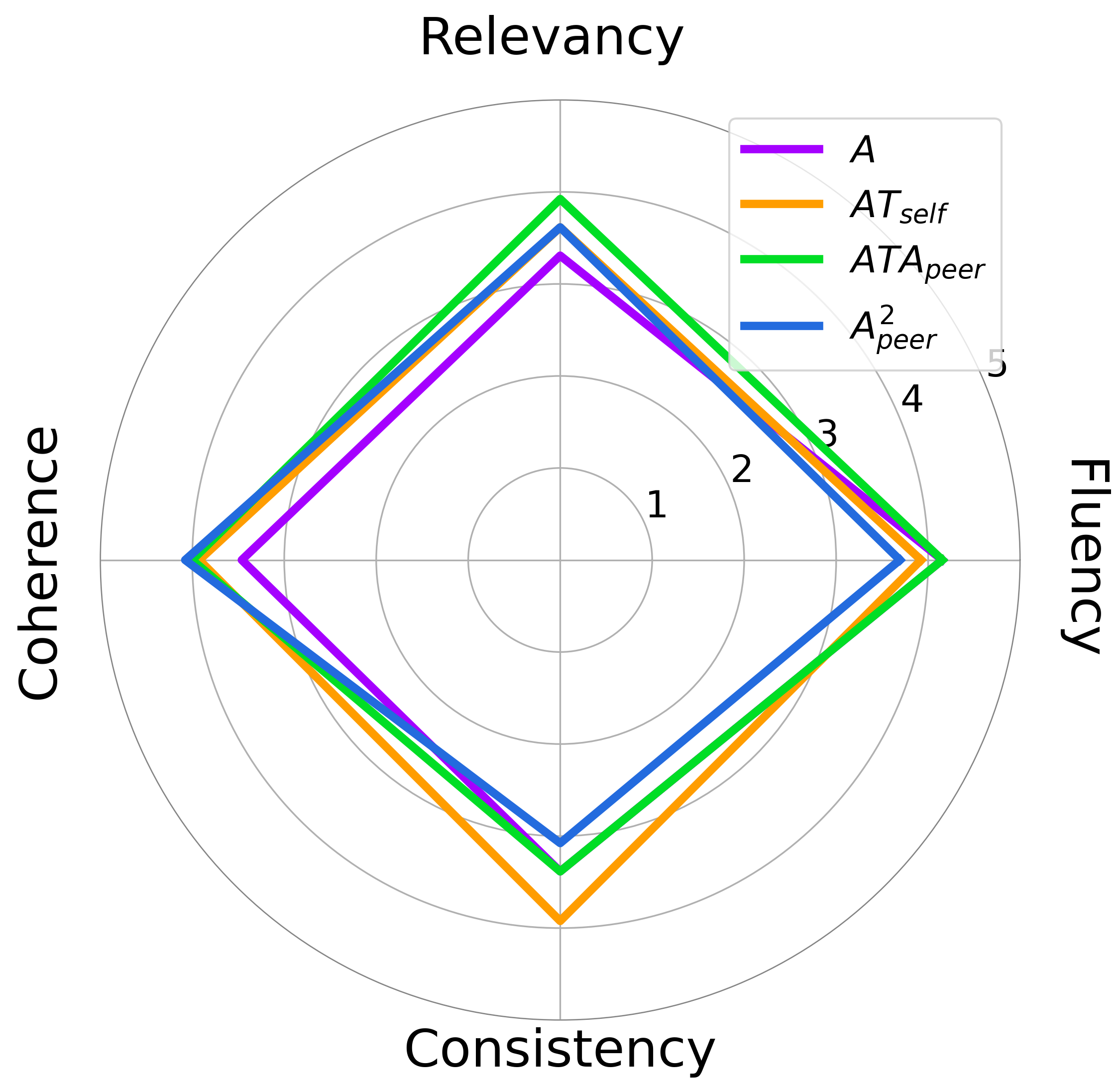}}
\quad
    \subfloat[\label{mt-simtask}]{%
    \includegraphics[trim=5 0 5 0, clip, width=0.33\textwidth]{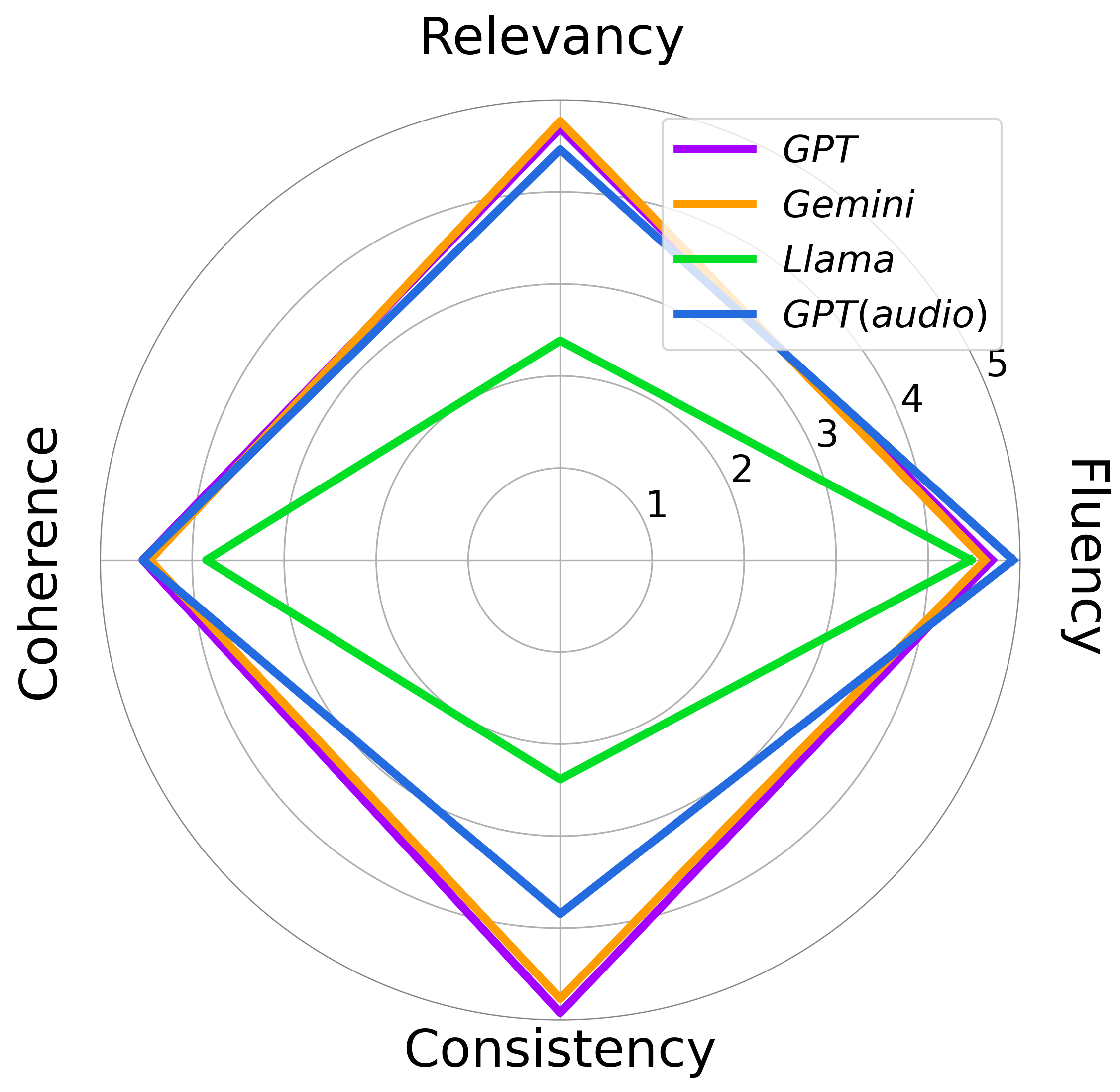}}\\
    \caption{\label{G-Eval} Spider charts of G-Eval Coherence, Relevancy, Fluency, and Consistency}
\end{figure*}

We generate LLM summaries for the 13 conversations with three long-context LLMs and one audio LLM. They serve as points of comparison for our human-written summaries and are used to examine our second hypothesis. We use Gpt-4o, Gemini-2.5-flash, Llama-3.2-1B-Instruct, and gpt-4o-audio-preview-2025-06-03 \citep{gpt4-techreport,grattafiori2024llama3herdmodels,comanici2025gemini25pushingfrontier}. These models were selected to fulfill a range of use cases and resources that might apply to researchers seeking to replicate this work: at the time of submission, gpt-4o is the gold-standard system of automatic abstractive summarization, gemini-2.5-flash is accessible from a budget standpoint as it can be used via the free tier of the Gemini platform, and Llama is an open-source model accessible from a hardware standpoint as it can be run on single, non-A100 GPUs. We include gpt-4o-audio-preview-2025-06-03 to have a reference representing audio LLMs, although the model is neither open-source nor accessible from a budget standpoint.

The text-based LLMs and the audio LLM were given a cleaned conversation transcript or audio recording, and they were prompted to generate a summary similar in length to human transcript-based summaries respecting the same summarization guidelines that were given to human annotators. 

\subsection{Annotator Tendencies}
We collected additional annotations of four CallHome conversations and two SWBDA conversations from eight of our annotators to assess individual annotator tendencies. The annotators wrote transcript-based summaries for three conversations and self-edited their summaries after listening to the audio. The same process was then repeated for three different conversations with the order of modalities reversed: annotators wrote an audio-based summary which they edited after reading the transcript. We analyze each annotator's summaries and calculate ROUGE-1/2/L, percentage of insertions and deletions, and BertScore between the initial and final summaries. We also calculate BertScore between the summary and the transcript, CREAM, and information density for each initial and final summary. We report similarity between annotators by calculating pairwise similarity between each set of two annotators for each metric as shown in Eq \ref{eq1}, where $x_1$ and $x_2$ are each annotator's summary's score for the given metric.  
\begin{equation} \label{eq1}
similarity = 1 - |x_1  - x_2| 
\end{equation}

We average each of the pairwise similarity values across the three conversations, and finally average across the entire set of annotators to obtain the similarity values in Table \ref{tab:Table 2}

\subsection{Data Analysis}
We analyzed the collected data from  several different perspectives: overall quality, lexical content, semantic content, length, and informativeness.\\
\textbf{Overall Quality:} We use G-Eval  with gpt-4o to evaluate each summary with respect to fluency, coherence, consistency, and relevance \citep{g-eval}. \\
\textbf{Editing Behavior:} Between initial and edited versions of summaries we calculate ROUGE-1/2/L , the percentage of insertions and the percentage of deletions \citep{rouge-score}. Higher ROUGE scores and fewer insertions and deletions between the initial and final summaries indicate less editing. We also calculate BertScore and BartScore between initial and edited summary versions to track how semantic content changes with editing \citep{Zhang2019BERTScoreET,BartScore-2021}.  \\
\textbf{Abstractiveness:} On every summary we find the percent of words in the summary which exist in the transcript (\% Lexical Overlap) and the percent of words in the summary which do not exist in the transcript (\% Novel Words). Summaries with lower lexical overlap or more novel words are more abstractive.
\\
\textbf{Semantic Content:} We compute BertScore and BartScore between each summary and its source's transcript. We also compute BertScore between the set of audio-based and transcript-based summaries. 
\\
\textbf{Length:} Compression ratio calculates the ratio of the number of words in the summary to the number of words in the transcript. 
\\
\textbf{Informativeness}: As we create and compare multiple sets of summaries with the intention of ranking the relative quality of the summaries, we require a strong reference-free metric to compare the summaries. We adopt a version of the Comparison-Based Reference-Free ELO-Ranked Automatic Evaluation for Meeting Summarization (CREAM) framework from \citet{CREAM-score} to assess the relative informativeness of each summary in comparison to the other summaries. We concatenate the summaries being compared into one paragraph and prompt an evaluating LLM to extract a list of up to 30 non-repetitive key facts from the paragraph. In a separate call, the key facts list is then provided with each of the individual summaries with instructions to indicate how many of the key facts in the list can be inferred from the summary. We refer to the ratio of supported key facts to the total number of key facts as the CREAM score, which we use as our relative measure of informativeness. We calculate CREAM scores on two sets of summaries: between all of the summaries (both human and LLM-written) and just between the human-written summaries. We also calculate information density as the ratio of each summary's CREAM scores to the number of words in the summary. 

We calculate paired t-test values between all 14 sets of combined human and LLM summaries for each metric to track statistical significance.

\section{Results}

\begin{table*}
\resizebox{\textwidth}{!}{%

\begin{tabular}{@{}cc|cccc|cccccc@{}}
\toprule

& & \multicolumn{4}{c|}{\textbf{Compared to initial summary}} & \multicolumn{5}{|c}{\textbf{Compared to the transcript}} \\
\cmidrule{3-7}\cmidrule{8-12}

\makecell{\textbf{Table} \\ \textbf{Row}} & \makecell{\textbf{Annotation} \\ \textbf{Workflow}} & \textbf{ROUGE-2} & \textbf{\%Insertions} & \textbf{\%Deletions} & \makecell{\textbf{Bert} \\ \textbf{Score}} & \makecell{\textbf{\%Lexical} \\ \textbf{Overlap}} & \makecell{\textbf{Bart} \\ \textbf{Score}}  & \makecell{\textbf{Bert} \\ \textbf{Score}} & \makecell{\textbf{CREAM} \\ \textbf{Score}} & \makecell{\textbf{Compression} \\ \textbf{Ratio}} & \makecell{\textbf{Info} \\ \textbf{Density}} \\
\midrule

1 & T & - & - & - & - & 0.60 & -6.88 & 0.80 & 0.38 & 0.55 & 0.0022 \\
2 & A & - & - & - & - & 0.68 & -6.88 & 0.80 & 0.19 & 0.27 & 0.0020 \\
\midrule
3 & TA$_{self}$ & 0.98 & 0.01 & 0.01 & 1.00 & 0.60 & -6.88 & 0.80 & 0.38 & 0.55 & 0.0022 \\
4 & AT$_{self}$ & 0.75 & 0.99 & 0.01 & 0.97 & 0.67 & -7.02 & 0.80 & 0.28 & 0.35 & 0.0026 \\
5 & TTA$_{peer}$ & 0.85 & 0.02 & 0.13 & 0.97 & 0.60 & -6.86 & 0.80 & 0.36 & 0.48 & 0.0024 \\
6 & ATA$_{peer}$ & 0.66 & 1.65 & 0.04 & 0.95 & 0.64 & -6.90 & 0.80 & 0.26 & 0.36 & 0.0022 \\
7 & T$^2_{peer}$ & 0.69 & 0.34 & 0.06 & 0.95 & 0.58 & -6.83 & 0.80 & 0.44 & 0.67 & 0.0020 \\
8 & A$^2_{peer}$ & 0.51 & 2.83 & 0.01 & 0.91 & 0.60 & -6.98 & 0.80 & 0.41 & 0.56 & 0.0023 \\
9 & T$^3_{peer}$ & 0.81 & 0.04 & 0.21 & 0.96 & 0.53 & -6.09 & 0.81 & 0.43 & 0.78 & 0.0015 \\
10 & A$^3_{peer}$ & 0.81 & 0.16 & 0.05 & 0.96 & 0.60 & -6.40 & 0.80 & 0.42 & 0.62 & 0.0021 \\
11 & T$^4_{peer}$ & 0.55 & 0.06 & 0.25 & 0.93 & 0.56 & -6.34 & 0.81 & 0.40 & 0.66 & 0.0016 \\
12 & A$^4_{peer}$ & 0.80 & 0.23 & 0.04 & 0.96 & 0.59 & -6.37 & 0.80 & 0.41 & 0.75 & 0.0017 \\
13 & T$^5_{peer}$ & 0.75 & 0.15 & 0.08 & 0.96 & 0.53 & -6.18 & 0.81 & 0.39 & 0.65 & 0.0016 \\
14 & A$^5_{peer}$ & 0.66 & 0.35 & 0.09 & 0.94 & 0.54 & -6.39 & 0.80 & 0.48 & 0.87 & 0.0023 \\
\midrule
15 & GPT & - & - & - & - & 0.55 & -6.89 & 0.80 & 0.51 & 0.63 & 0.0025 \\
16 & Gemini & - & - & - & - & 0.53 & -6.94 & 0.80 & 0.64 & 0.59 & 0.0034 \\
17 & Llama & - & - & - & - & 0.56 & -6.86 & 0.81 & 0.30 & 0.60 & 0.0015 \\
18 & GPT (Audio) & - & - & - & - & 0.54 & -6.98 & 0.81 & 0.59 & 0.65 & 0.0027 \\
\bottomrule
\end{tabular}}
\caption{Summaries-to-summaries and Summaries-to-Transcript comparison metrics.
}
\label{tab:Table 1}
\end{table*}

\subsection{Audio vs Text Summaries}
We find that un-edited audio-based summaries are shorter, less informative, and less abstractive than un-edited transcript-based summaries. This is signified by lower CREAM scores, higher percentages of lexical overlap and lower percentages of novel words in row 2 of Table \ref{tab:Table 1}.  Our findings regarding audio-based versus transcript-based summaries align with those of \citet{Sharma:2024:SpeechvsText}, who also found that audio-based summaries are shorter, less informative, and less abstractive than transcript-based summaries. There is, however, no significant difference in information density between transcript-based and audio-based summaries. In this context, and in the rest of this section of results, we use the word significance to refer to statistical significance.  

Audio-based summaries also receive significantly smaller relevancy scores than transcript-based summaries, shown by the spider plots (a) and (b) in Figure \ref{G-Eval} showing G-Eval scores for transcript-based summaries and their edited versions and audio-based summaries and their edited versions respectively. However, we believe that these low relevancy scores may be due to length and the somewhat subjective nature of summarization: the information captured in the short length of an audio-based summary may not include details that the model deemed to be the focal points of the conversation. Our metric for assessing informativeness, CREAM, is more forgiving in this sense as it narrows the scope of potential focal points by extracting key facts from the concatenation of summaries. We use CREAM to calculate information density, and we observe audio-based and transcript-based summaries have similar information density, similar semantic similarity to the conversation transcript, and high semantic similarity to each other (BertScore = 0.87). Thus,  we conclude that audio-based summaries are equally focused and relevant to the transcript as transcript-based summaries, despite being shorter in length and containing less information. 

\begin{figure}[!ht]
\begin{center}
\includegraphics[width=\columnwidth]{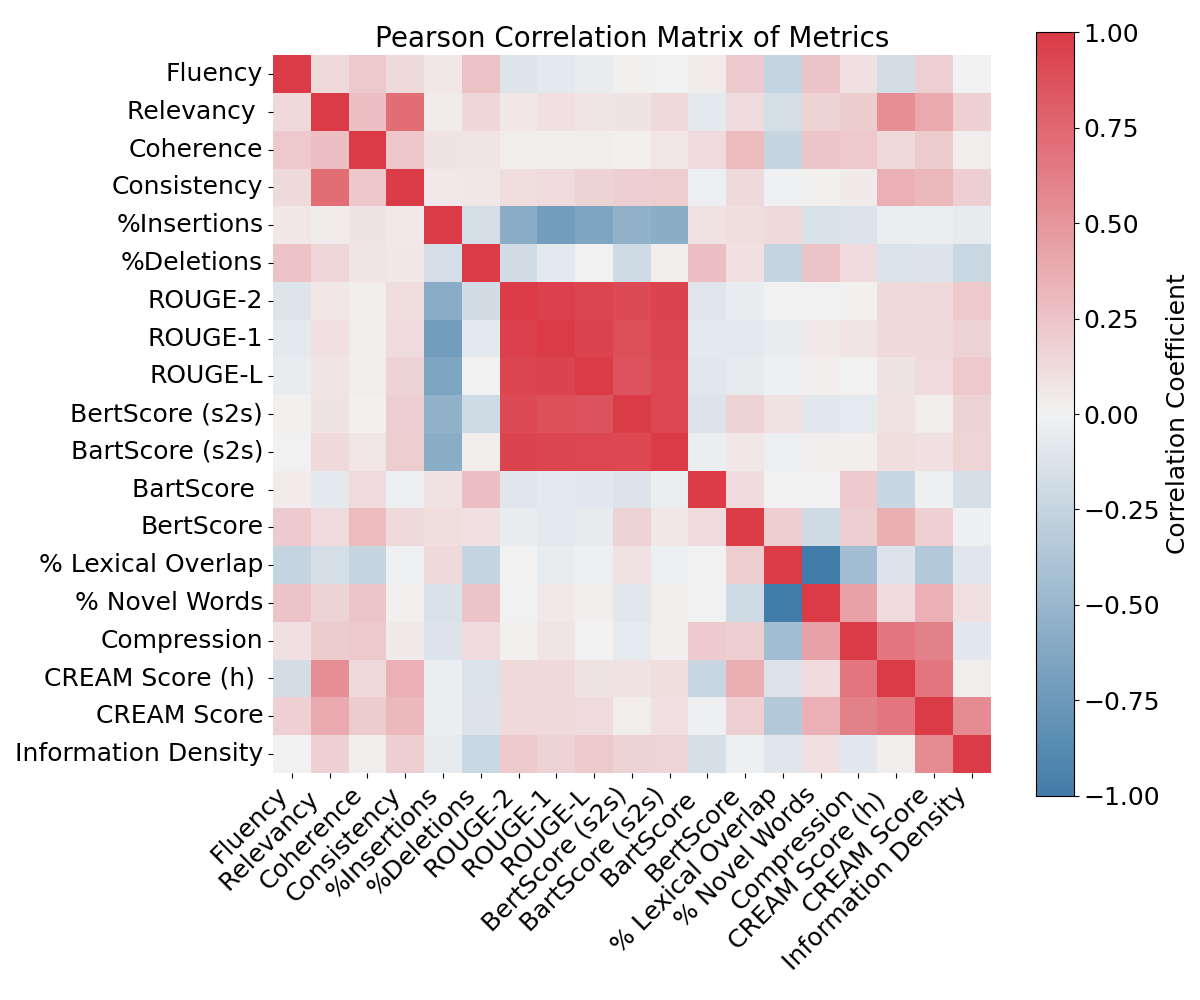}
\caption{Heatmap of the Pearson correlation between all of the different metrics calculated. BertScore (s2s) and BartScore (s2s) correspond to the respective metrics calculated between initial and edited versions of summaries, whereas BertScore and BartScore refer to the metrics calculated between a summary and the source transcript.}
\label{fig.4}
\end{center}
\end{figure}

Table \ref{tab:Table 1} shows a subset of the full set of metrics calculated. We show a subset because many metrics (ie. ROUGE-1 and ROUGE-2) are highly correlated. The most informative metrics in Table \ref{tab:Table 1} and show the full heatmap of Pearson correlation coefficients in Figure \ref{fig.4}. 

We note in Figure \ref{fig.4} that while BertScore and BartScore between initial and final summaries are highly correlated, BertScore and BartScore between the individual summary and the transcript are not. We attribute this to the fact that  summaries are written using third-person narrative while the conversation transcripts contain exchanges from two people speaking in the first person. As BartScore reports the log-likelihood of a candidate text given a reference text, this translation between perspectives likely leads to substantially lower BertScores even when the summary is relevant to the content of the transcript.

\subsection{Self vs Peer Editing}
Excluding cases of consecutive peer-editing, we find that self-editing (rows 3 and 4 in Table \ref{tab:Table 1}) produces significantly less changes to the base summary compared to any peer-editing method (rows 5 and 5 in Table \ref{tab:Table 1}). We note that the difference between edits made to the base summary between ATA$_{peer}$ and AT$_{self}$ are not significant.

In rows 7, 9, 11, and 12, iterative peer-editing using only the transcript T$^{2-5}_{peer}$, we neither see significant differences in the amount of edits made over multiple rounds nor do we see significant changes in informativeness, information density, or the quality dimensions of G-Eval. 

However, in iterative peer-editing with audio, the first editing round (row 8) produces significantly more edits than any of the later rounds (rows 10, 13, and 14), and that extensive editing results in a significantly more informative A$^2_{peer}$ summary than the original audio-based summary. We believe this happens because the terseness of the starting audio-based summary allows peer-editors to identify a lot of missing information to include. There are no significant changes in informativeness or information density between A$^2_{peer}$ (row 8) and A$^{3-5}_{peer}$ (rows 10, 12, 14) summaries, indicating that after the first round of editing the informational content of the summary stabilizes. We also note that there is no significant difference in informativeness between A$^{2-5}_{peer}$ (rows 8, 10, 12, 14) and their T$^{2-5}_{peer}$ counterparts in rows (7, 9, 11, 13). This shows that one round of peer-editing overcomes the differences in informativeness between audio-based and transcript-based summaries.

\subsection{How do the final summaries compare?}
We see that iterative peer-editing overcomes differences in informativeness associated with the base summary's input modality: there are no significant differences in CREAM score between T$^{2-5}_{peer}$ (rows 7, 9, 11, and 13) and A$^{2-5}_{peer}$  summaries (rows 8, 10, 12, and 14) in Table \ref{tab:Table 1}. On the other hand, AT$_{self}$ (row 4) and ATA$_{peer}$ summaries (row 6) are significantly less informative and more compressed, shown by lower CREAM scores and greater compression ratios in comparison to TA$_{self}$ (row 3) and TTA$_{peer}$ summaries (row 5) respectively. Therefore, self-editing and peer-editing with both the audio and the transcript maintain the differences in informativeness that exist between audio-based and transcript-based summaries, while peer-editing with only one modality of information overcomes these differences. This may be an indication of the cognitive load of the two tasks: asking an annotator to peer-edit a summary using both the transcript and the audio as a reference may overwhelm them with information, leading to less comprehensive editing than when they only use the audio as a reference. 

We also see that while  AT$_{self}$ summaries have similar levels of lexical overlap with the transcript as the audio-based summaries, A${^2_{peer}}$ and ATA$_{peer}$ summaries have significantly less lexical overlap with the transcript. This shows that peer editing with audio is an effective method of increasing abstractiveness in summaries. 

We observe no major differences in semantic similarity across base summary modality or editing style of the final summaries, showing that all styles of annotation produced summaries that were similarly relevant to the semantic content of the transcripts. We also see lower consistency scores across the board when A is the base summary, indicating that annotators incorporate information from the audio into their summaries which does not exist in the transcript. 

\subsubsection{How do LLM-generated summaries compare to the human-written summaries?}

We look to informativeness to compare the relative quality of the different summaries. While the best performing LLMs are more informative than the base summaries and the summaries that have been produced through a single round of editing, there is no significant difference in informativeness between either $T^{2-5}_{peer}$ or A$^{3-5}_{peer}$ and any of the LLM summaries (rows 15 - 18 in Table \ref{tab:Table 1}). 

We observe that the Gemini model produces the most informative summaries, the Llama model produces least informative summaries, and summaries from Gpt-4o and Gpt-4o Audio rank between the Gemini and Llama models. 

Summaries written with Llama (row 17) are only significantly more informative than AT$_{self}$, ATA$_{peer}$, and the original audio-based summaries (rows 4, 6, and 2 respectively). Gpt-4o (row 15) and Gpt-4o Audio (row 18) summaries are not significantly more informative than T$^{2-5}_{peer}$ and A$^{3-5}_{peer}$ summaries. And Gemini summaries (row 16) are not significantly more informative than T$^{5}_{peer}$ and A$^{3-5}_{peer}$.

This shows that multiple rounds of human peer-editing with transcript or audio can produce summaries that are as informative as our best performing LLMs.



\subsection{Analysis of Annotator Tendencies}

We find high average similarity across annotator editing behavior. We also find that annotators tend to write summaries of similar length, informativeness levels, and  semantic similarity to the transcript. These results are shown in Table \ref{tab:Table 2}.

\begin{table}
\begin{center}
\begin{tabularx}{\columnwidth}{l cc}

      \toprule
      \textbf{Metric} & \textbf{T} & \textbf{TA$_{self}$}  \\
        \midrule
        ROUGE-2 & - & 0.77\\
        \% Insertions & - & 0.81\\
        \% Deletions & - & 0.78\\
         BertScore (s2s) & - & 0.82\\
      \% Lexical Overlap & 0.76 & 0.76\\
        Compression Ratio & 0.67 & 0.67\\

        BertScore & 0.83 & 0.83 \\
        BartScore & 0.56 & 0.56 \\
        CREAM & 0.67 & 0.68 \\
        Info Density & 0.83 & 0.83 \\      
      \bottomrule
\end{tabularx}
\caption{Average pairwise similarity between annotators for three T and TA$_{self}$ summaries. Larger values show greater similarity, with 1 representing identical tendencies.}
\label{tab:Table 2}
\end{center}
\end{table}

\section{Conclusion}
This study contributed a comparison of workflows to collect human summaries from speech. We identified two potential tradeoffs in using humans to summarize speech: humans may write less informative summaries from audio than from transcripts, and humans may write summaries that are less informative than LLM summaries.  We  explored ten human summarization workflows and compared the resulting summaries to those from four LLMs to examine these potential tradeoffs. Our findings confirmed that un-edited summaries written from audio are less informative and more compressed than those written from transcripts. They also show that iterative peer-editing overcomes the differences between audio and transcript based summaries, and audio-based summaries iteratively peer-edited with the audio are as informative as their text based counterparts \textit{and} as informative as LLM summaries. 

These findings have direct and significant implications for data collection. Our validated workflow proves that high-quality, prosody-informed speech datasets can be collected even in situations where transcripts are not available. This is an important finding because ASR systems still perform poorly in certain domains, and it is expensive to manually transcribe audio. Sole requirement of the audio modality will allow these domains to be included in speech summarization much more easily than if high-quality transcripts were also required. 

Since iterative peer-editing of audio summaries produces a product on par with LLMs, the summarization benchmarks created using this method will be both diverse and rigorous. These results also make a case to explore human-machine teaming where the initial summary is written by LLM and then edited by a human as future work. 

We arrive at the conclusions of this paper through metrics that make comparisons against the source transcript (e.g., BertScore, Compression Ratio) or comparisons across the collected summaries (e.g., CREAM). Notably, we do not make any comparisons between the summary and the source audio. 

While we assume that human summaries, when created by an annotator exposed to the audio, include some form of prosodic information, we have no metrics with which to validate any of this information. For the same reason, we also cannot examine how the human-written summaries represent the audio in relation to the GPT-4o Audio preview summaries. We leave the development of a metric assessing how accurately a summary represents the prosodic contents of speech as important future work. 


\section{Limitations}

Our study's conclusions are subject to several limitations associated with annotator availability. 

Firstly, we limit our analysis to summaries generated from 13 conversations as this was the maximum number of summaries we could collect during the period of time for which our annotators were hired. The data for this study was collected over a total of 6 weeks with each annotator spending three to five hours a week on annotation. We did not require annotators to complete a minimum number of summaries per hour to avoid influencing the quality of the written and edited summaries. While the consistency of our findings across ten workflows suggests robustness, a larger corpus would strengthen the generalizability of the peer-editing effect across diverse conversational styles.

Secondly, constraints around annotator availability necessitated using two distinct sets of annotators for summary collection. The initial set produced the baseline summaries (Figure \ref{fig.2}, rows 1-3), while a second set, responsible for the iterative editing and the final baseline audio summary (Figure \ref{fig.2}, last row; Figure \ref{fig.3}), received an expedited training protocol. Furthermore, due to limitations in annotator availability we did not collect data to analyze similarity across the second set of annotators as we prioritized collecting edited versions of the 13 conversations.

\nocite{*}
\section{Bibliographical References}\label{sec:reference}

\bibliographystyle{lrec2026-natbib}
\bibliography{less_authors}


\newpage             
\onecolumn          
\appendix
\section{Conversation IDs}

\begin{table}[h]
    \centering
    \caption{CallHome Conversations Used and Their Word Counts}
    \label{tab:example}
    \begin{tabular}{ll}
        \toprule
        \makecell{\textbf{CallHome} \\ \textbf{Conversation ID}}
 & \makecell{\textbf{Transcript}\\ \textbf{ Word Count}} \\
        \midrule
        en\_4184 & 884 \\
        en\_4415 & 1012 \\
        en\_4145 & 1032 \\
        en\_4628 & 970 \\
        en\_4665 & 1037 \\
        en\_4335 & 1219 \\
        en\_4576 & 956 \\
        en\_4660 & 1062 \\
        en\_4112 & 696 \\
        \bottomrule
    \end{tabular}
\end{table}

\begin{table}[h]
    \centering
    \caption{Switchboard Conversations Used and Their Word Counts}
    \label{tab:example}
    \begin{tabular}{ll}
        \toprule
        \makecell{\textbf{Switchboard} \\ \textbf{Conversation ID}}
 & \makecell{\textbf{Transcript}\\ \textbf{ Word Count}} \\
        \midrule
        sw03182 & 765 \\
        sw02012 & 1876 \\
        sw02366 & 1621 \\
        sw03181 & 897 \\
        \bottomrule
    \end{tabular}
\end{table}

\section{LLM Prompting}
\begin{center}
    \captionof{figure}{System and user prompts used for G-Eval}
    \label{fig:task1_prompts}
\end{center}

\begin{promptbox}
\textbf{G-Eval System Prompt:} \\
You will be given one summary of a conversation. Your task is to rate the summary on one metric. Please make sure you read and understand these instructions carefully. Please keep this document open while reviewing, and refer to it as needed.
\end{promptbox}

\begin{promptbox}
\textbf{G-Eval User Fluency Prompt:} \\
Evaluation Criteria: \\Fluency (1-5) - Fluency refers to how grammatically correct, natural, and easy to read the text is. Fluency encompasses grammar, punctuation, spelling, word choice, and phrasing. Do not consider factual correctness or logical flow. Do not penalize the summary for using gender neutral pronouns to refer to people or for using names to refer to people instead of pronouns. Follow these steps carefully: \\Step-by-step analysis: Analyze the text sentence by sentence. Point out any grammatical errors, awkward phrasing, or unnatural expressions. Also note any strengths, such as clear and elegant sentences. \\Summary of analysis: Summarize your detailed observations in one paragraph, explaining the overall fluency quality of the summary. \\Numeric score: At the end of your response, provide a final fluency score from 1 to 5 using this scale: \\1 — Very poor fluency (many serious grammatical errors, very awkward or unnatural) \\2 — Poor fluency (several grammatical errors or awkward phrases, difficult to read) \\3 — Fair fluency (some noticeable errors, but understandable overall) \\4 — Good fluency (minor errors or slightly awkward phrasing, generally natural) \\5 — Excellent fluency (no significant errors, reads perfectly naturally) \\Report your score as\\'Final Score: [score]' \\Summary to evaluate:\\"
\end{promptbox}

\begin{promptbox}
\textbf{G-Eval User Coherence Prompt:} \\
Evaluation Criteria: \\ Coherence (1-5) - Coherence refers to how well the sentences and ideas in the text logically connect to each other and how the information flows from beginning to end. It does not consider grammatical correctness or factual accuracy. Follow these steps carefully:\\  Step-by-step analysis: Analyze the summary sentence by sentence. Identify whether each sentence logically follows from the previous one, whether there are abrupt topic shifts or jumps, and whether the overall structure makes sense. Highlight any strengths, such as smooth transitions or a strong overall structure. \\Summary of analysis: Summarize your detailed observations in one paragraph, explaining the overall coherence quality of the summary.\\Numeric score: At the end of your response, provide a final coherence score from 1 to 5 using this scale:\\1 — Very poor coherence (very disjointed, no logical flow, very hard to follow)\\2 — Poor coherence (frequent logical jumps, lacks clear structure)\\3 — Fair coherence (some jumps or slightly confusing sections, but generally follows a rough order)\\4 — Good coherence (well organized, smooth flow with minor issues)\\5 — Excellent coherence (very logical and easy to follow, no noticeable issues)\\ Report your score as\\'Final Score: [score]' \\ Summary to evaluate:\\
\end{promptbox}

\begin{promptbox}
\textbf{G-Eval User Consistency Prompt:}

Evaluation Criteria:\\Consistency (1-5) - Consistency refers to the factual alignment between the summary and the source document. The summary should contain only statements that are entailed by the source document. You are given a conversation transcript (the source document) and you must assess factual consistency of the provided summary with respect to the transcript. The conversation transcript includes two speakers, A and B, and speaker turns are indicated by A: and B: . Follow these steps carefully: \\ Step-by-step analysis: Analyze the summary statement by statement. Validate whether the claim made in each statement is supported by the source document. \\Summary of analysis: Summarize your detailed observations on the summary's consistency in one paragraph. \\Numeric score: At the end of your response, provide a final consistency score from 1 to 5 using this scale:\\1 - Very poor consistency (many claims the summary are completely unsupported by the source document) \\2  - Poor consistency (claims in the summary are frequently unsupported by the source document) \\3 - Fair consistency (some claims in the summary are unsupported by the source document) \\4 - Good consistency (nearly all of the summary claims are supported by the source doucment) \\5 - Excellent consistency (every single claim in the summary is supported by the source document) \\Report your numeric score as\\'Final Score: [score]' \\ Source document: \\ Summary to evaluate:
\end{promptbox}

\begin{promptbox}
\textbf{G-Eval User Relevancy Prompt:}

Evaluation Criteria: Relevance (1-5) - Relevance refers to the selection of information in the summary and if it captures the most important content from the source document. A summary should include only the important information from the source document. You are given a conversation transcript (the source document) and you must assess factual consistency of the provided summary with respect to the transcript. The conversation transcript includes two speakers, A and B, and speaker turns are indicated by A: and B: . Follow these steps carefully: \\ Step-by-step analysis: Go through the summary and identify each distinct piece of information that is included. For each identified piece of information, assess whether it is an important part of the source document. \\Summary of analysis: Summarize your detailed observations on the summary's relevancy in one paragraph. \\Numeric score: At the end of your response, provie a final relevancy score from 1 to 5 using this scale:\\1 - Very poor relevancy (much of the information included in the summary is irrelevant to the most important details of the source document)\\2 - Poor relevancy (information in the summary is frequency irrelevant to the most important details of the source document) \\3 - Fair relevancy (some information in the summary is irrelevant to the most important details of the source document)\\4 - Good relevancy (nearly all of the information in the summary is relevant to the most important information in the source document)\\5 - Excellent relevancy (all information in the summary is relevant to the most important information in the source document) \\Report your numeric score as\\'Final Score: [score]' \\ Source document:  \\ Summary to evaluate: 
\end{promptbox}
 
\begin{center}
    \captionof{figure}{System and user prompts to produce summaries from Gemini-2.5-Flash, GPT-4o, GPT-4o Audio preview, and Llama-3.2-1B-Instruct }
    \label{fig:task1_prompts}
\end{center}

\begin{promptbox}
\textbf{GPT-4o Audio preview Prompt:}\\
The included transcript and audio are from a phone call between two people. The conversations may:

  - Begin mid-conversation or with participants setting up the recording
  - End abruptly before the conversation concludes
  - Contain segments in a foreign language, denoted in the transcript as: <a: xyz>, where <a> specifies the foreign language (e.g., Spanish) and <xyz> represents the spoken phrase.

  Your task is to succinctly summarize the conversation so that someone who does not have access to the transcript or the audio can understand the main points discussed. This means your summary must be: 
  
  - Accurate: Only include information that can be backed up by the transcript or audio
  
  - Coherent: Easy to read. No overly long sentences, overly terse, or awkward sentence structure 
  
  - Concise and free from redundancies

  - Self Contained: No references to information from the conversation that the reader does not have access to

  Your Summary Should Include:

  1. High Level Information

  \quad- Capture the main points and most important details of the conversation that will help someone reading your summary understand the context of the call and what was discussed. Avoid capturing small details. 

  \quad- For example, if one speaker mentions that the sky is clear outside, you should not include it in your summary unless it ties in to a main point of the conversation ie. if the conversation is about a drought in that speaker's town.  

  2. Speaker References

  \quad- Use **Person A** and **Person B** initially to refer to the speakers.
  \quad- If a speaker's name is mentioned, you may use it but clearly link it:  
    e.g., “Person A, whose name is Selma, described..."

  3. Sufficient Context

  \quad- Avoid summaries that assume the reader knows the speakers or their relationship.
  \quad- Prefer:  
    “According to Person A, someone they know named Sue returned home last week”  
    Over:  
    “Sue came home last week”

  4. Topic Shifts

  \quad- When the conversation topics shift suddenly, use transitional phrasing so that the reader can understand how that information relates to the rest of the summary:  
    ex.  “The conversation then shifted to…”  
        “They abruptly changed topics to discuss…”

  5. Logical Connections

  \quad- The logical connection between all the details included in your summary should be clear. Logical connections can be things like cause and effect, agreement/ disagreement, sequential events, etc. Once someone has read your entire summary they should be able to explain how each piece of information included in the summary connects to the rest of the summary. 

  \quad- Information that does not have a clear logical connection to the rest of the summary (and is not the result of a sudden shift in topic indicated through transitional wording) should be omitted. 

  6. Summarization instead of Paraphrasing

  \quad Summarizing means condensing the conversation to its essential points while preserving meaning. Paraphrasing means restating each part of the conversation in slightly different language — this is not what we want. 

  \quad\quad - Summarization (this is what we want)
      “Person A asked for advice about a job offer, and Person B suggested prioritizing work-life balance.”

  \quad\quad - Paraphrasing (this is NOT what we want)  
      “Person A talked about a new opportunity. Person B replied with a story about their past job. Then Person A asked a question. Then Person B...”

\end{promptbox}

\begin{promptbox}
\textbf{Gemini-2.5-Flash Prompt:}\\
The transcript below captures a phone call between two individuals. These conversations may:

\quad -Start mid-discussion or as participants set up the recording.

\quad -End abruptly before the conversation concludes.

\quad -Contain segments in a foreign language, marked like this: <a: xyz>, where <a> specifies the foreign language (e.g., Spanish) and <xyz> represents the spoken phrase.

Your task is to produce an accurate, highly concise summary of the key points in this conversation, strictly adhering to the following requirements:

\quad Word-Count Constraint:

\quad\quad- The summary must be no more than 50\% of the transcript's length in word count. For example: If the transcript is 400 words, the summary must be no longer than 200 words. If the transcript is 300 words, the summary must be no longer than 150 words.
Use strategic omissions of minor or redundant details to achieve this level of compression while capturing the essential takeaways.

Summary Requirements:

\quad Focus on Key Takeaways:

\quad\quad - Condense the conversation into its core ideas and most significant details.
Exclude minor, non-essential details, repetitions, or elaborations that do not directly add to the reader’s understanding of the main discussion.

\quad Strong Summarization Over Paraphrasing:

\quad\quad - Summarization Goal: “Person A asked for advice on switching careers, and Person B emphasized the importance of maintaining work-life balance while outlining potential risks.”

\quad\quad - Avoid Paraphrasing: “Person A said they were thinking about switching careers and listed reasons for their hesitation. Then, Person B shared a story about their career path before offering advice on work-life balance…”

\quad Logical and Coherent:

\quad\quad - Ensure the summary is well-organized, addressing the flow of topics and transitions:

\quad\quad\quad Example: "The discussion began with Person A’s job concerns before shifting to personal finances."

\quad\quad - Signal changes explicitly (e.g., "The conversation then shifted to…").

\quad Self-Contained:

\quad\quad - Avoid relying on any external knowledge or context beyond what is directly stated in the transcript.

\quad Accurate Speaker Attribution:

\quad\quad - Use “Person A” and “Person B” consistently, unless names are provided, in which case you may clarify their identity once (e.g., "Person B, whose name is Mark…").
Revision Focus:

Your absolute priority is to compress the text into a concise form:

\quad - Prioritize brevity while preserving accuracy and readability.

\quad - Ensure the summary does not exceed 50\% of the transcript’s length.

\quad - Avoid redundancy, overly long sentences, or minor details that dilute the main points.

Reminder:\\
Summarize smartly: Focus on the main takeaways, simplify structure, and stay adherent to the 50\% word limit of the transcript.
\end{promptbox}

\begin{promptbox}
\textbf{GPT-4o Prompt:}\\
 The below transcript is from a phone call between two people. These conversations may:

  \quad - Begin mid-conversation or with participants setting up the recording
  
  \quad - End abruptly before the conversation concludes
  
  \quad - Contain segments in a foreign language, noted in the transcript as:  
    , where `a` is the language tag (e.g., `spanish`) and `xyz` is the spoken phrase

  Your task is to summarize the conversation so that someone who does not have access to the transcript or the audio can understand the main points discussed. This requires your summaries to be: 
  
 \quad - Accurate: Only include information that can be backed up by the transcript or audio
  
  \quad - Coherent: Easy to read. No overly long sentences or awkward sentence structure 
  
  \quad - Concise and free from redundancies

  \quad - Self Contained: No references to information from the conversation that the reader does not have access to

  Your Summary Should Include:

  1. High Level Information

  \quad - Capture the main points and most important details of the conversation that will help someone reading your summary understand the context of the call and what was discussed. Avoid capturing small details. 

  \quad - For example, if one speaker mentions that the sky is clear outside, you should not include it in your summary unless it ties in to a main point of the conversation ie. if the conversation is about a drought in that speaker's town.  

  2. Speaker References

  \quad - Use **Person A** and **Person B** initially to refer to the speakers.
  \quad - If a speaker's name is mentioned, you may use it but clearly link it:  
    e.g., “Person A, whose name is Selma, described..."

  3. Sufficient Context

  \quad - Avoid summaries that assume the reader knows the speakers or their relationship.
  \quad - Prefer:  
    “According to Person A, someone they know named Sue returned home last week”  
    Over:  
    “Sue came home last week”

  4. Topic Shifts

  \quad - When the conversation topics shift suddenly, use transitional phrasing so that the reader can understand how that information relates to the rest of the summary:  
    ex.  “The conversation then shifted to…”  
        “They abruptly changed topics to discuss…”

  5. Logical Connections

  \quad - The logical connection between all the details included in your summary should be clear. Logical connections can be things like cause and effect, agreement/ disagreement, sequential events, etc. Once someone has read your entire summary they should be able to explain how each piece of information included in the summary connects to the rest of the summary. 

  \quad - Information that does not have a clear logical connection to the rest of the summary (and is not the result of a sudden shift in topic indicated through transitional wording) should be omitted. 

  6. Summarization instead of Paraphrasing

  \quad - Summarizing means condensing the conversation to its essential points while preserving meaning. Paraphrasing means restating each part of the conversation in slightly different language — this is not what we want. 

      \quad\quad Summarization (this is what we want): 
      
      \quad\quad\quad “Person A asked for advice about a job offer, and Person B suggested prioritizing work-life balance.”

      \quad\quad Paraphrasing (this is NOT what we want): 
      
      \quad\quad\quad “Person A talked about a new opportunity. Person B replied with a story about their past job. Then Person A asked a question. Then Person B...”
\end{promptbox}

\begin{promptbox}
\textbf{Llama-3.2-1B-Instruct:}

Below is the transcript of a phone call between two individuals. These conversations may:

\quad Begin mid-discussion or with participants setting up the recording.
\quad End abruptly before the conversation concludes.
\quad Feature segments in a foreign language, denoted as <a: xyz>, where a specifies the foreign language (e.g., Spanish) and xyz represents the spoken phrase.

Your task is to generate a brief summary of the conversation so someone without access to the transcript can fully understand its scope. Your summary should:

\quad - Include only information supported by the transcript.

\quad - Avoid assumptions, speculation, or omitting key details.

\quad - Contain approximately half the word count of the transcript. For example, a 400 word transcript should receive a ~200 word summary

\quad - Weave the main points of the conversation into the summary 

\quad - Arrange ideas logically with smooth transitions between topics and speakers.

\quad - Clearly identify shifts in conversation topics: e.g., “After discussing budget concerns, the discussion shifted to vacation planning…”

\quad -Refer to speakers as Person A and Person B. If names are provided, clarify them: e.g., “Person A, whose name is Selma, mentioned…”

\quad -Be able to stand alone, without requiring external knowledge from the transcript.

Do not explicitly list out the key points of the conversation. 

\end{promptbox}

\begin{center}
    \captionof{figure}{System and user prompts used for CREAM}
    \label{fig:task1_prompts}
\end{center}

\begin{promptbox}
\textbf{System Prompt for Obtaining Key Facts List:}

You are an assistant designed to process key facts from provided text.

\textbf{User Prompt for Obtaining Key Facts List:}
You will be provided with a paragraph. Your task is to decompose the paragraph into a set of \"key facts\". A \"key fact\" is a single fact written as briefly and clearly as possible, encompassing at most 2-3 entities. There should not be any repeated \"key fact\". Here are nine examples of key facts to illustrate the desired level of granularity (please be careful that these are examples, and do NOT use them as actual key facts):

\quad * Kevin Carr set off on his journey from Haytor. 

\quad * Kevin Carr set off on his journey from Dartmoor.

\quad * Kevin Carr set off on his journey in July 2013.

\quad * Kevin Carr is less than 24 hours away from completing his trip.

\quad * Kevin Carr ran around the world unsupported.

\quad * Kevin Carr ran with his tent.

\quad * Kevin Carr is set to break the previous record.

\quad * Kevin Carr is set to break the record by 24 hours.

\quad * The previous record was held by an Australian. 

Instruction: 

First, read the paragraph carefully. Second, decompose the paragraph into (at most 30) key facts. Provide your answer in JSON format. The answer should be a list of dictionaries whose keys are "key fact": ["key fact": "first key fact", "key fact": "second key fact", "key fact": "third key fact"] 
\end{promptbox}

\begin{promptbox}
\textbf{System Prompt to Ascertain the Number of Key Facts Supported by a Single Summary:}

You are an assistant designed to validate key facts based on provided text.

\textbf{User Prompt to Ascertain the Number of Key Facts Supported by a Single Summary:} 

You have a set of key facts and a summary. Your task is to assess if each key fact can be inferred from the summary.

Instruction: First, compare each key fact with the summary. Second, check if that key fact is inferred from the summary, and for each key fact respond "Yes" if it can be inferred from the summary or "No" if it cannot be inferred from the summary. If "Yes", specify the line number(s) of the summary sentence(s) relevant to each key fact. Provide your answer in JSON format. The answer should be a list of dictionaries whose keys are "key fact", "response", and "line number": ["key fact": "first key fact", "response": "Yes", "line number": [1], "key fact": "second key fact", "response": "No", "line number": [], "key fact": "third key fact", "response": "Yes", "line number": [1, 2, 3]]

\end{promptbox}

\end{document}